\documentclass[letterpaper, 10 pt, conference]{ieeeconf}  

\IEEEoverridecommandlockouts                              

\usepackage{bm,amsmath,amssymb}
\usepackage{graphicx}
\graphicspath{ {images/} }
\usepackage{tikz} \usetikzlibrary{positioning}
\usepackage{hyperref}
\usepackage{subcaption}
\usepackage[font=footnotesize]{caption}

\usepackage{diagbox}

\usepackage{array}
\newcolumntype{P}[1]{>{\centering\arraybackslash}p{#1}}
\newcolumntype{M}[1]{>{\centering\arraybackslash}m{#1}}

\DeclareMathOperator*{\argmin}{arg\,min}

\def\Arrow{\makebox[6.26399pt]{$\blacktriangleright$}~}

\newcommand{\vect}[1]{\boldsymbol{#1}}

\title{\LARGE \bf
Contact Transfer: A Direct, User-Driven Method for Human to Robot Transfer of Grasps and Manipulations
}

\author{Arjun Lakshmipathy$^{1}$ Dominik Bauer$^{2}$ Cornelia Bauer$^{2}$ Nancy S. Pollard$^{1,2}$
\thanks{$^{1}$Computer Science Department, Carnegie Mellon University}%
\thanks{$^{2}$Robotics Institute, Carnegie Mellon University}%
}

\begin{document}

\maketitle
\thispagestyle{empty}
\pagestyle{empty}

\begin{abstract}

We present a novel method for the direct transfer of grasps and manipulations between objects and hands through utilization of contact areas. Our method fully preserves contact shapes, and in contrast to existing techniques, is not dependent on grasp families, requires no model training or grasp sampling, makes no assumptions about manipulator morphology or kinematics, and allows user control over both transfer parameters and solution optimization. Despite these accommodations, we show that our method is capable of synthesizing kinematically-feasible whole hand poses in seconds even for poor initializations or hard-to-reach contacts. We additionally highlight the method's benefits in both response to design alterations as well as fast approximation over in-hand manipulation sequences. Finally, we demonstrate a solution generated by our method on a physical, custom-designed prosthetic hand.

\end{abstract}

\section{INTRODUCTION}

Transfer of human grasps and manipulation demonstrations to robot hands has been a long standing and challenging problem in the robotics community. Though numerous anthropomorphic hands have been developed (e.g., \cite{allegrohand, shadowhand, diftler2002robonaut, miahand, larahand,bauer2020design}), no current manipulator yet matches the dexterous capabilities of the human hand. In addition, simpler hands are often desired based on considerations such as weight, cost, and ease of manufacture and control.

Towards this end, many research efforts have endeavored to develop methods for robustly transferring dexterous capabilities to robot hands with characteristics that may differ from the hand of the human demonstrator. The majority of existing methods fall into pure learning, hand pose tracking and retargeting, or a hybrid of both. Pure reinforcement learning methods have produced convincing results even for complex manipulations \cite{rajeswaran2018drl, zhu2019dextrousrl, andrychowicz2020learning}, but occasionally yield policies with unexpected or undesirable behaviors due to the abstract nature of the reward function. Tracking methods instead capture the movements of both the human hand and the object being manipulated with the intention of grounding the learned policy \cite{jia2020visionbased} or directly transferring the motion to the target manipulator \cite{antotsiou2018taskretargeting, garciahernando2020physicspose}; unfortunately, occlusion and noise during the tracking process, as well as morphological differences between the human and target hand, complicate the process. Summarily, it is worth noting that these techniques focus on matching observations in hand or object space as opposed to the interaction interface between the two bodies.

Contacts instead present an alternate and often complementary means of encoding grasps and manipulations. Historical works have used contact information to prune kinematically-infeasible solutions \cite{pollard1997parallelgrasps}, match object and hand shapes \cite{li2007shape, hillenbrand2012contactwarping}, and drive physics based reasoning \cite{ye2012contactsampling}. Recent works exploit entire contact regions to synthesize grasps for morphologically diverse manipulators \cite{brahmbhatt2019contactgrasp}, optimize coarse poses \cite{grady2021contactopt}, and help train pose generation models \cite{jiang2021contactconsistency}. The catch, however, is that most existing techniques incur one of three possible drawbacks: single point approximation \cite{pollard1997parallelgrasps, li2007shape, hillenbrand2012contactwarping, ye2012contactsampling}, strong dependence on grasp dependent hand shape priors \cite{grady2021contactopt, jiang2021contactconsistency}, or expensive computation due to exhaustive grasp sampling \cite{brahmbhatt2019contactgrasp}. 

\begin{figure}[t]
\centering
\includegraphics[width=0.9\linewidth]{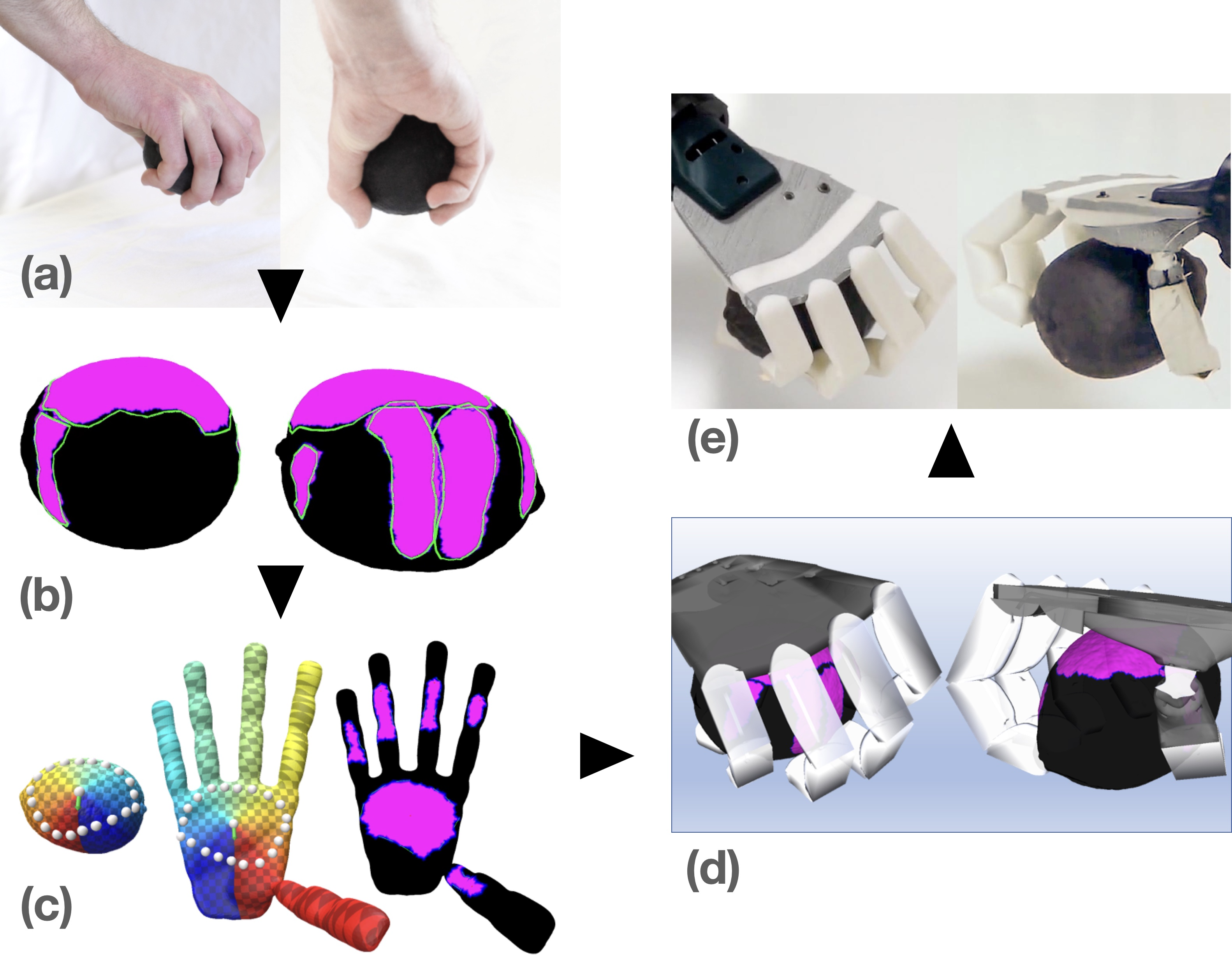}
\caption{High level overview of our framework. (a) Human demonstrations of grasps and manipulations are collected and (b) reconstructed using existing techniques. (c) Contact regions are then transferred from object to target manipulator, and (d) the whole hand kinematic pose is computed from the transferred contacts via optimization. (e) The robot hand then utilizes the final articulated solution.}
\vspace*{-0.2in}
\label{fig:overview}
\end{figure}

In response to these drawbacks, we present a geometric framework for the \textit{direct}, intuitive, and rapid transfer of contacts observed and collected from objects to morphologically diverse hands that fully preserves contact shape and is not dependent on external hand pose datasets or grasp sampling. More specifically, this paper makes the following contributions:

\begin{enumerate}
    \item a novel approach to \textit{directly} transfer contact areas of arbitrary shape to morphologically diverse manipulators in a fast and fully customizable manner
    \item a cheap optimization strategy to quickly synthesize kinematically-feasible grasps and in-hand manipulations by exploiting correspondences generated by the transfer process
\end{enumerate}

Our approach is illustrated in Figure~\ref{fig:overview}. We demonstrate that our method can accommodate hands with high degrees of freedom and contact areas of widely varying shapes. We also show that our framework produces solutions that are robust to poor initial conditions as well as difficult-to-reach contacts. The plug-and-play capability awarded by our transfer method makes it especially well suited for manipulator prototyping, which we illustrate in concluding experiments. Finally, we demonstrate a synthesized grasp solution on a physical, custom-designed prosthetic hand.

\section{RELATED WORK}

Numerous prior efforts have targeted the problem of recovering and optimizing manipulator pose from contacts, the majority of which can be broadly classified under either single point or area based approaches.

Single point methods use individual object points as the basis for extracting the corresponding hand configuration. Mapping individual object and hand contact points allows for the use of traditional inverse kinematics \cite{pollard1997parallelgrasps, qiu2021IKhighdim}; however, in grasps that contain a large number of contacts, the problem is generally overconstrained and may instead be best solved using an optimization approach where the user can specify the importance of different tradeoffs (e.g., match contacts, match hand pose, etc.). Optimization approaches can additionally incorporate kinematic constraints and endeavor to find least-squares \cite{qiu2021IKhighdim} or force-closure inducing \cite{rosales2012prehensile, ye2012contactsampling} solutions. But while these approaches are fast and straightforward, single point contacts are an idealized representation of real-world situations. Additionally, approximating contact areas as single points sacrifices useful information.

Consequently, some recent research efforts have turned instead to utilizing contact regions. Though a number of existing works have considered contacts purely through geometric reasoning, including independent contact regions \cite{roa2009icr3d, krug2010icrclosure} and directly mapping individual links to mesh slices \cite{rosales2011contactregions}, recent data driven works have produced high fidelity contact maps by instead capturing contacts directly from human demonstrations \cite{brahmbhatt2019contactdb, brahmbhatt2020contactpose} or synthesizing them from vision based retargeting and simulation \cite{jiang2021contactconsistency}. These maps have subsequently been utilized to synthesize grasps for morphologically diverse manipulators \cite{brahmbhatt2019contactgrasp} as well as optimize coarse estimates \cite{grady2021contactopt, jiang2021contactconsistency}; however, owing to the difficulty of mapping contact regions to articulated manipulators, these methods are dependent on priors or exhaustive grasp sampling, which thus limit their robustness and speed. 

This paper directly addresses the mapping problem by proposing a method to transfer object contact areas to any manipulator in a manner that fully preserves relative contact point distances regardless of underlying geometry, making it possible to combine the benefits of single point and area based techniques. We  show that the resulting contact area correspondence enables the use of a simple, yet robust optimization to compute the best matching, kinematically-feasible hand pose. Our technique does not require sampling or external datasets and can be used on any hand design, and is intentionally designed to allow full user control over both the transfer and optimization process, which we show leads to fast, robust, and qualitatively reasonable results for both static grasps and in-hand manipulations. We demonstrate results for a range of simulated robot hands, a graphical human hand, and a real custom-designed prosthetic hand.

\section{PROCEDURE}

Our algorithm consists of several stages, roughly illustrated in Figure \ref{fig:overview}. First, thermal traces of contact patches generated from human grasps and manipulations are captured and reconstructed onto object meshes using techniques from prior works \cite{brahmbhatt2019contactdb, lakshmipathy2021contacttracing}. The process outlined in this paper then transfers contact patches from object surfaces onto a ``skin" of the manipulator, and subsequently projects contacts from the skin onto the articulated hand. Finally, we utilize the mapping to compute grasps and manipulations with the target manipulator that best match the correspondence. Each stage is described in detail in the proceeding subsections.

\subsection{Contact Capture, Reconstruction, and Tracing}

Contact regions resulting from human grasps are first captured using a combination of thermochromic spray painted objects and RGBD imaging, and subsequently mapped onto object mesh surfaces using a reconstruction process detailed in existing works \cite{brahmbhatt2019contactdb, lakshmipathy2021contacttracing}. Regions are then annotated and processed into \textit{patches}, which are represented by a single \textit{root} ($PR$) mesh vertex and collection of \textit{boundary} ($PB$) mesh vertices. Boundaries are then down-sampled to produce \textit{interpolation boundary} ($IB$) sets of 20-30 mesh vertices, which are then traced from initial to final positions in the case of manipulations. Detailed definitions and process descriptions are available in \cite{lakshmipathy2021contacttracing}.

\subsection{Skinning and Projection}

\begin{figure}
\centering
\includegraphics[width=0.9\linewidth,height=0.45\linewidth]{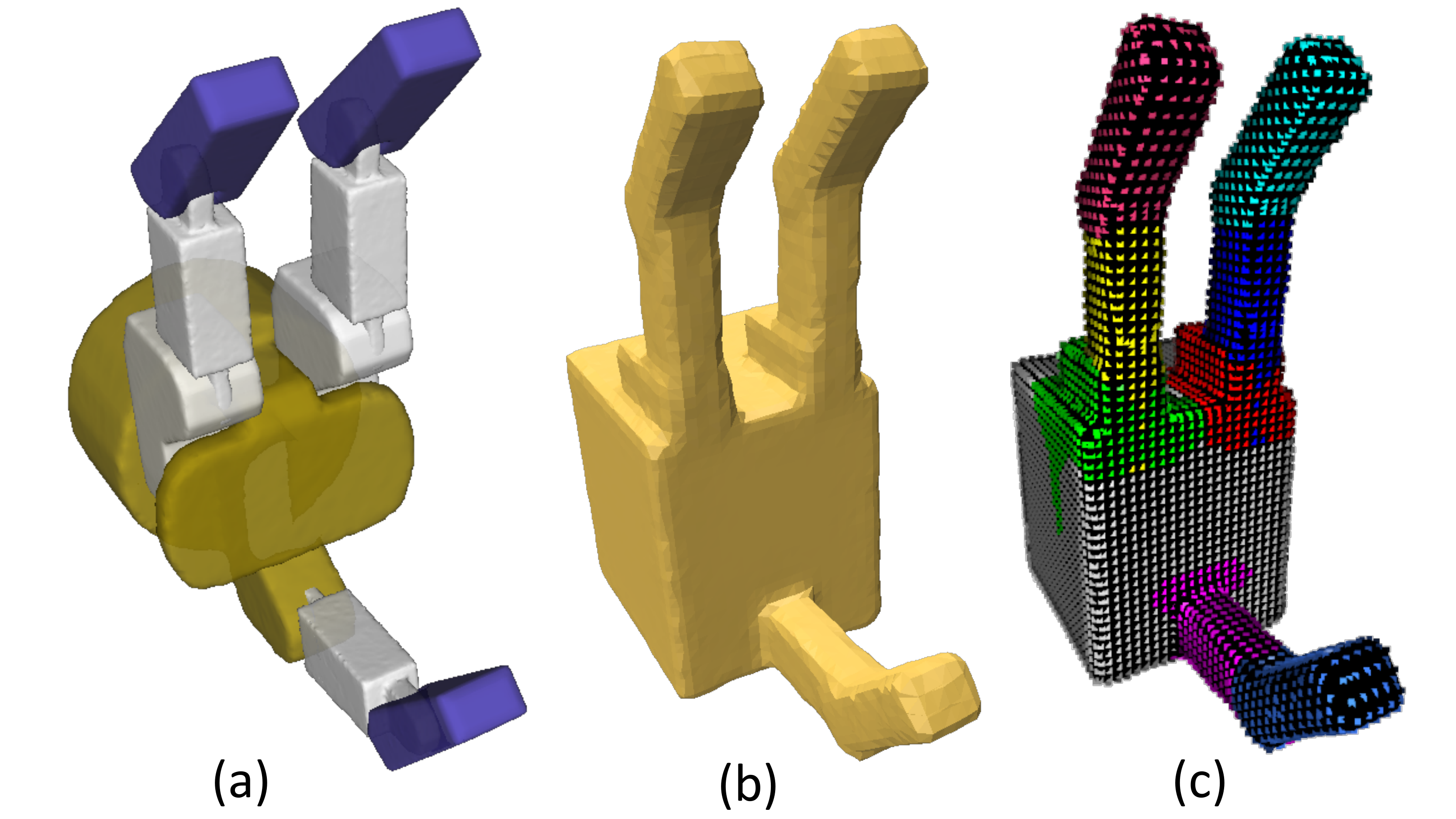}
\caption{The (a) articulated Barrett Hand can be (b) approximately skinned by stitching together primitive collision geometries. (c) The projection process partitions the skin into groups based on the link locations of the articulated hand.}
\label{fig:skinBarrett}
\end{figure}

Geometry processing algorithms are typically designed to operate over connected surfaces; however, the large majority of robotic systems are articulated, comprised instead of independent surface links connected by joints. These surfaces may also exhibit degenerate characteristics, including non-manifold edges, obtuse triangles, and near-zero area faces, which, combined with articulation, render many geometry processing algorithms useless.

Therefore, we instead construct a fully manifold ``skin" of each manipulator and then project vertices from the skin back on to the original articulated geometry using KD tree \cite{bentley1975kdtree} approximation. Figure \ref{fig:skinBarrett} illustrates the skinning and projection process. In particular, we note that skins can be constructed from even coarse mesh approximations such as primitive collision geometries. Post projection, points on the skin which have no nearby articulated vertex below threshold $\epsilon$ are ignored during optimization. We use Blender \cite{blender} with the Phobos attachment to construct skins and the Open3d library \cite{zhou2018open3d} to perform the projection.

\subsection{Contact Patch Transfer}

\begin{figure}
\centering
\includegraphics[width=0.9\linewidth,height=0.5\linewidth]{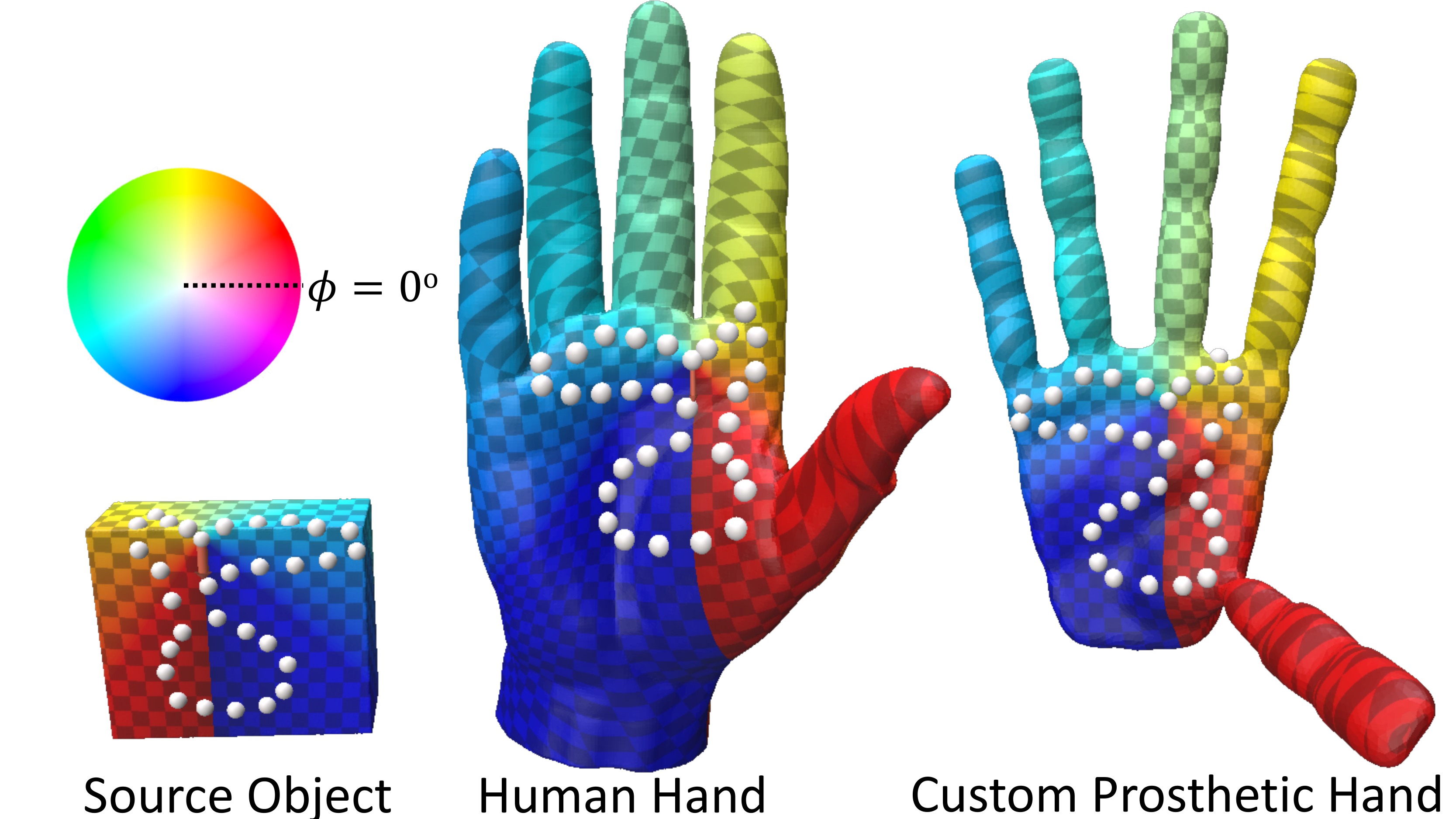}
\caption{Illustration of our logmap-based contact patch transfer process. Accurate relative distances and angles at all mesh vertices enable transfer of even large, irregularly shaped patches (outlined by white dots) to widely differing hand geometries.}
\label{fig:contactTransfer}
\end{figure}

The key idea of our transfer process is that patch shapes on the object and manipulator are equivalent, regardless of the surface on to which they are projected. More formally, the distance and direction of the $IB$ points from the root should be preserved under a geometry independent discrete logarithmic map $\mathbf{L}_{r, \phi}$ \cite{schmidt2006dem, sharp2019vhm}. We therefore require four parameters, all of which are visually adjustable by the user: the patch's root vertex on the hand skin $\mathbf{v}_{r,h}$ and object $\mathbf{v}_{r,o}$, as well as the tangent vector direction from the hand $\vec{\mathbf{v}}_{r,h}$ and object $\vec{\mathbf{v}}_{r,o}$ in which to begin the sweep. The corresponding $IB$ set on the hand can then be entirely computed as:

\begin{equation}
    \begin{array}{rrclcl}
        \displaystyle \mathbf{v}^*_{IB_i,h} = \argmin_{\mathbf{v}_h} & \multicolumn{3}{l}{\|\mathbf{L}_{r, \phi}(\mathbf{v}_{h}) - \mathbf{L}_{r, \phi}(\mathbf{v}_{o})\ ||_2^2}\\
        \textrm{s.t.} & \mathbf{L}_{r, \phi}(\mathbf{v}_{h}) &=& f(\mathbf{v}_{r,h}; \vec{\mathbf{v}}_{r,h})\\
         & \mathbf{L}_{r, \phi}(\mathbf{v}_{o}) &=& f(\mathbf{v}_{r,o}; \vec{\mathbf{v}}_{r,o})\\
    \end{array}
    \label{eq:logmapmatch}
\end{equation}

Finding these correspondences thus requires computation of logmap coordinates $(r, \phi)$ for all vertices on both object and skin. Naively, these quantities can be computed by tracing geodesics from the root to all vertices ($r$) and propagating the root tangent vector via parallel transport ($\phi$) \cite{guggenheimer2012differential}; unfortunately, these operations are notoriously expensive. The Vector Heat Method proposed by Sharp et. al. \cite{sharp2019vhm}, however, reduces to seconds the parallel transport of $both$ divergence and root tangent vectors through the use of short time heat diffusion. As Figure \ref{fig:contactTransfer} illustrates, this method can successfully transfer patches of arbitrary shape on to widely differing surface geometries, can be performed in seconds, and guarantees both relative distance and angle preservation. We use the Polyscope viewer \cite{polyscope} and Geometry Central library \cite{geometrycentral} to enable user selection of $(\mathbf{v}_{r,h}; \mathbf{v}_{r,o}; \vec{\mathbf{v}}_{r,h};\vec{\mathbf{v}}_{r,o})$, solve Eq. \ref{eq:logmapmatch}, and export the final mapping.

\subsection{Extension to Manipulations}

We note that the aforementioned procedure can only transfer static grasps; however, manipulations require dynamic contacts capable of moving, appearing, or disappearing from the hand at any time. To account for this behavior, we first transfer initial and final grasps and then adopt the procedure of Lakshmipathy et. al. \cite{lakshmipathy2021contacttracing} to evolve patches on the hand in conjunction with evolution of patches on the object.

\subsection{Optimization Procedure}

\begin{figure*}
\centering
\subcaptionbox{}{\includegraphics[width=0.224\linewidth,height=0.125\linewidth]{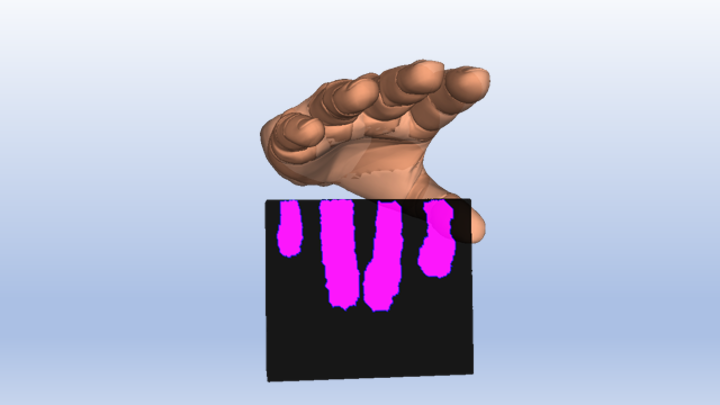}}
\Arrow
\subcaptionbox{}{\includegraphics[width=0.224\linewidth,height=0.125\linewidth]{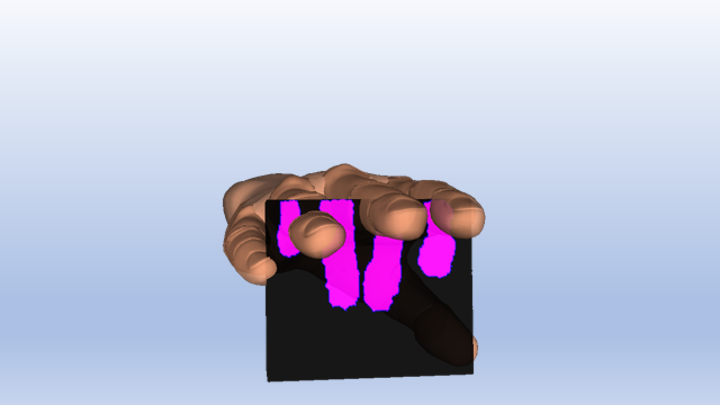}}
\Arrow
\subcaptionbox{}{\includegraphics[width=0.224\linewidth,height=0.125\linewidth]{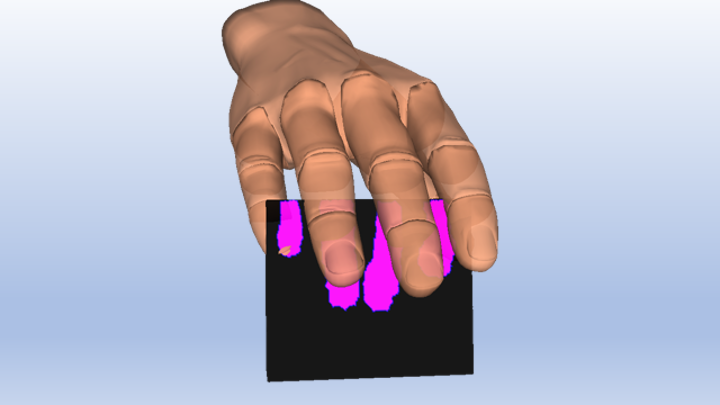}}
\Arrow
\subcaptionbox{}{\includegraphics[width=0.224\linewidth,height=0.125\linewidth]{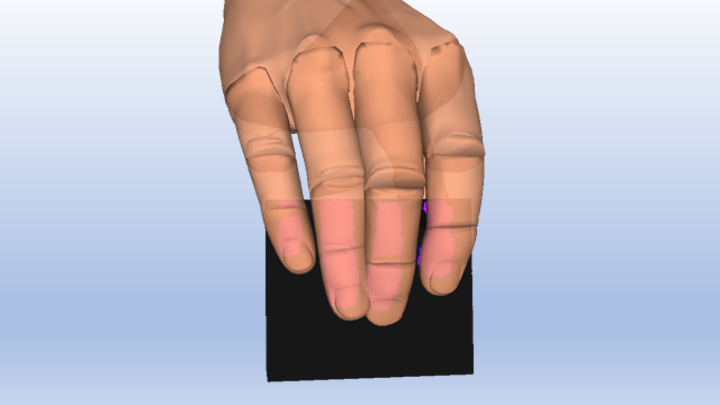}}
\caption{Optimization progress for finding a feasible box power grasp at (a) 0, (b) 250, (c) 800, and (d) 1,000 iterations.}
\label{fig:optimization}
\end{figure*}

Computing the associated $IB$ mappings between the object and hand enabled use of a straightforward, three-term optimization problem to determine the best kinematic hand pose $\vect{\theta}^*$:

\begin{equation}
    \begin{array}{rrclcl}
        \displaystyle \vect{\theta}^* = \argmin_{\vect{\theta}} & \multicolumn{3}{l}{\sum_{i}^{M}  (\lambda_d \Gamma^D_{i} + \lambda_n \Gamma^N_{i}) + \sum_{j}^{J} \lambda_p \Gamma^P_{j}}\\
        \mathrm{s.t.} & \vect{\theta}_L \leq \vect{\theta} \leq \vect{\theta}_U
    \end{array}
    \label{eq:opt}
\end{equation}

where $\vect{\theta}$ is the degree of freedom vector, $M$ is the total number of corresponding contact vertices, $J = |\vect{\theta}|$, $\vect{\theta}_L$ and $\vect{\theta}_U$ are the lower and upper bounds of each degree of freedom respectively, $\Gamma^D_{i}$, $\Gamma^N_{i}$, and $\Gamma^P_{j}$ are the distance, normal, and prior pose deviation penalty terms for each corresponding pair of points $i$ and joint $j$ respectively, and $\lambda_d$, $\lambda_n$, and $\lambda_p$ are weighting hyperparameters. Each penalty term is elaborated upon in the proceeding paragraphs.

First, we introduce $\Gamma_{D,i}$ to minimize the $L_2$ distance between each pair $i$ of corresponding hand and object contacts:

\begin{equation}
    \Gamma_{D,i} = ||\mathbf{p}_{o,i}-\mathbf{p}_{h,i}(\vect{\theta})||_2^2
\end{equation}

and $\Gamma_{N,i}$ to encourage anti-alignment of vertex normals:

\begin{equation}
    \Gamma_{N,i} = (1 + \mathbf{n}_{h,i}(\vect{\theta}) \cdot \mathbf{n}_{o,i})^2
\end{equation}

where each pair of object and hand contact points are denoted by $o,i$ and $h,i$ respectively, and hand point locations as well as vertex normal orientations are determined by the current hand pose $\vect{\theta}$.

We additionally introduce a third term to penalize deviation from $\vect{\theta}_P$, a hand pose prior:

\begin{equation}
    \Gamma_{P,i} = ||\vect{\theta} - \vect{\theta}_P||_2^2
\end{equation}

The prior is initialized in the same way for all examples and serves as a user control for customization. At the start of the optimization, $\vect{\theta}_P$ is set to the default pose, thereby penalizing rest pose deviation. During subsequent calls, $\vect{\theta}_P$ is set to the optimal solution from the last set of iterations; however, if the user edits the default guess (e.g. moves a finger, drags the palm, etc.), $\vect{\theta}_P$ is instead set to the user edited pose. As a result, $\Gamma_{P,i}$ evolves as the optimization proceeds to reflect current progress and user direction.

It is worth noting that Eq. \ref{eq:opt} does not contain a collision penalty term as suggested by several prior works \cite{brahmbhatt2019contactgrasp, meixner2019evolutionary, hazard2018automated}. Omission of this term is intentional and ultimately led to both substantial drops in solution discovery time as well as improvements in the proposal of solutions for difficult contact maps (see Figure \ref{fig:hardGrasp}). 

The fully differentiable nature and cheap computation cost of Eq. \ref{eq:opt} allows utilization of fast, gradient based solvers even using numerical gradient approximation; however, we empirically found that different solvers provided slightly different solutions. We ultimately settled on the Method of Moving Asympototes \cite{svanberg2002mma} for generating the majority of our results, but advocate implementing multiple solvers and allowing the user to choose between results. Each method was capped at 1,000 iterations per optimization call to intermittently update $\vect{\theta}_P$ and allow the user to edit intermediate estimates using IK. Figure \ref{fig:optimization} illustrates successive guesses over the course of a single optimization call. We used the NLOpt library \cite{johnson2017nlopt} for defining and solving Eq. \ref{eq:opt}.

\section{RESULTS}

Similar to Brahmbhatt et. al. \cite{brahmbhatt2019contactgrasp}, we performed a series of experiments using four different manipulators - a 20 degree of freedom human hand \cite{miller2004graspit}, an Allegro hand \cite{allegrohand}, a Barrett hand \cite{barretthand}, and a custom-made anthropomorphic prosthetic hand - across several static grasps and manipulations of common household items. Items and contact maps were borrowed from several existing datasets, including ContactDB \cite{brahmbhatt2019contactdb}, Contact Tracing \cite{lakshmipathy2021contacttracing}, and YCB \cite{calli2015ycb}. The primary desiderata were speed and scalability, robustness, and flexibility to accommodate design or intermediate optimization modifications. Only results for qualitatively reasonable grasps are reported. All simulations were conducted using the Dynamic Animation and Robotics Toolkit (DART) \cite{lee2018dart}.

\subsection{Grasps and Manipulations}

\begin{figure*}
\centering
\includegraphics[width=\linewidth,height=0.415\linewidth]{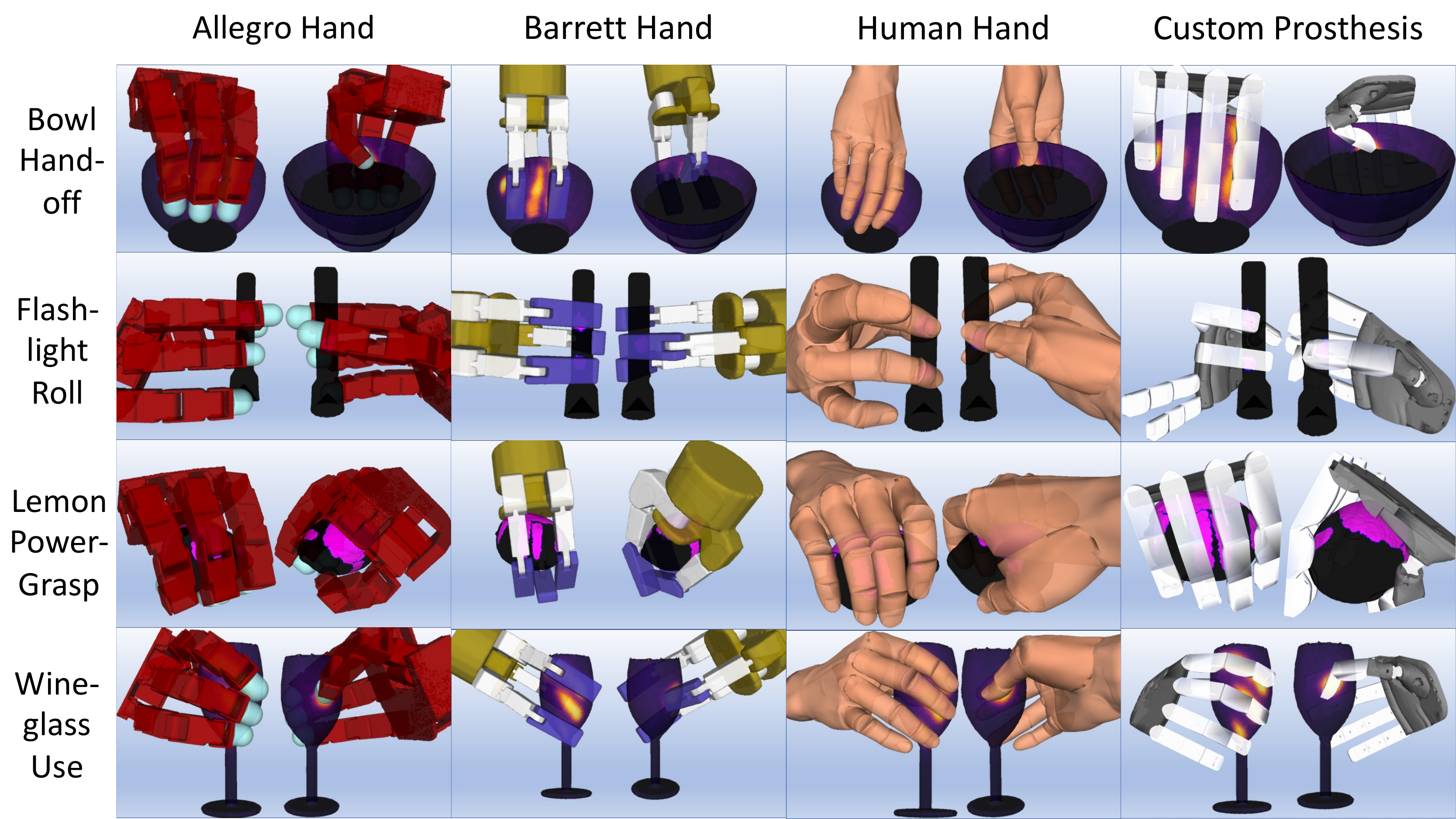}
\caption{Grasps synthesized as a result of our transfer procedure across a variety of kinematically diverse manipulators, objects, and grasps.}
\label{fig:graspMontage}
\end{figure*}

\begin{figure*}
\centering
\includegraphics[width=0.224\linewidth,height=0.12\linewidth]{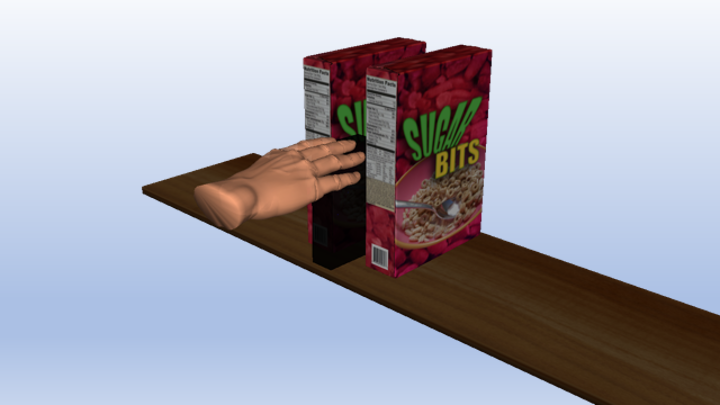}
\Arrow
\includegraphics[width=0.224\linewidth,height=0.12\linewidth]{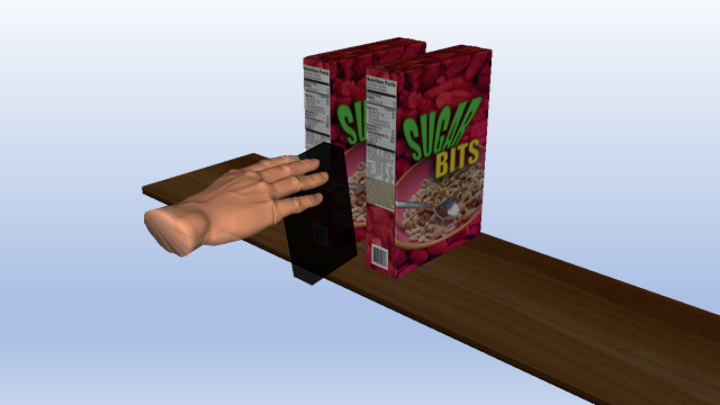}
\Arrow
\includegraphics[width=0.224\linewidth,height=0.12\linewidth]{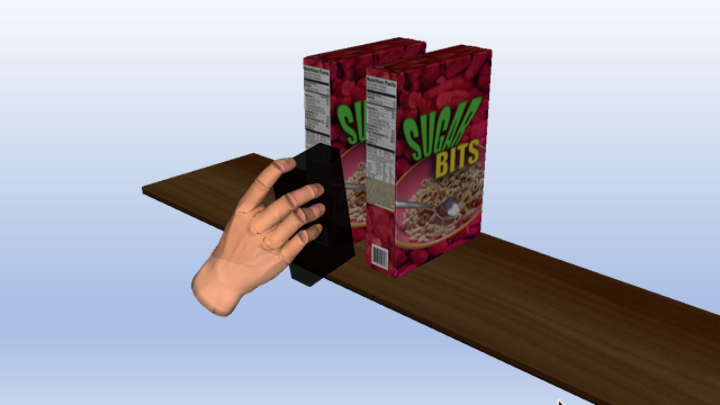}
\Arrow
\includegraphics[width=0.224\linewidth,height=0.12\linewidth]{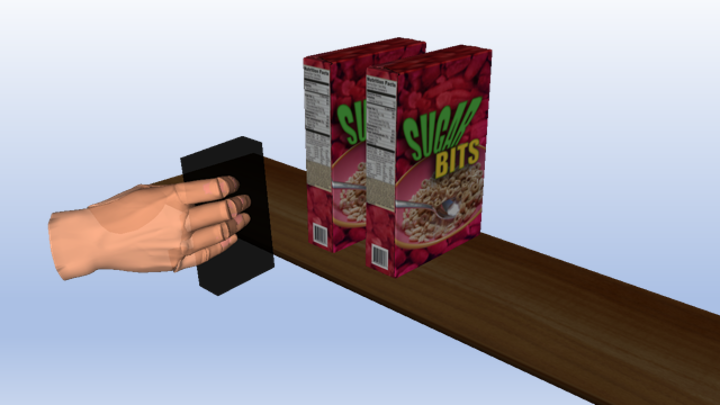}
\includegraphics[width=0.224\linewidth,height=0.12\linewidth]{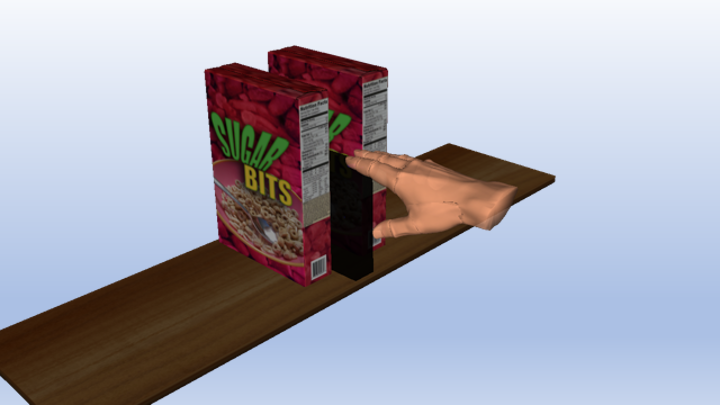}
\Arrow
\includegraphics[width=0.224\linewidth,height=0.12\linewidth]{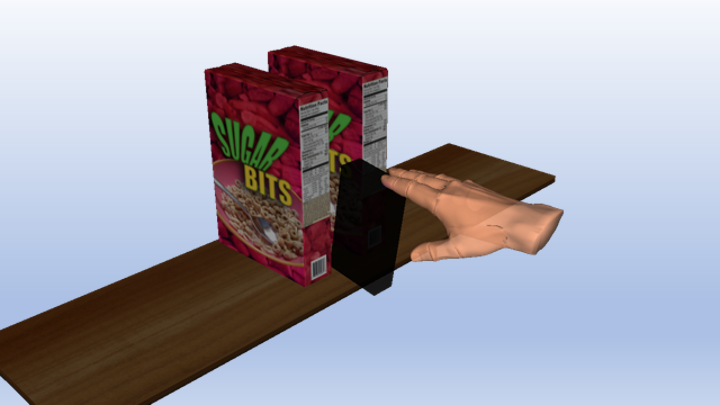}
\Arrow
\includegraphics[width=0.224\linewidth,height=0.12\linewidth]{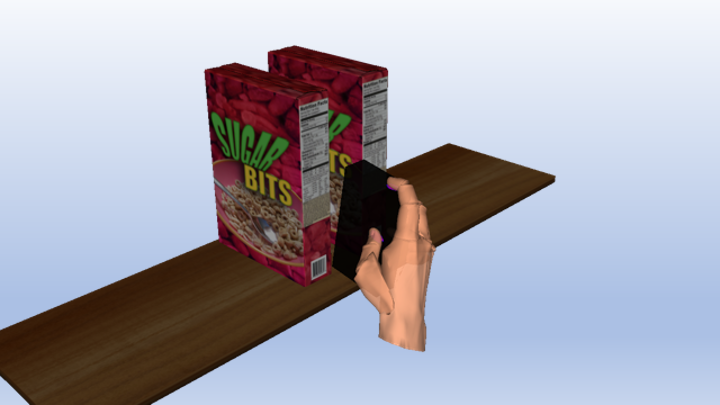}
\Arrow
\includegraphics[width=0.224\linewidth,height=0.12\linewidth]{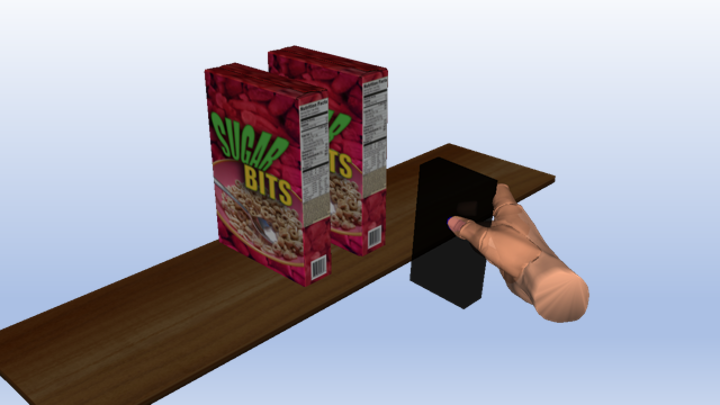}
\caption{Demonstration of using the method to quickly approximate a kinematic grasp time series spanning a full simple manipulation of pulling a box of sugar off a kitchen shelf. Poses for successive manipulation steps can be generated at real-time speeds due to the inclusion $\vect{\theta}_p$ and contact evolution time series data.}
\label{fig:fullManipulation}
\end{figure*}

Figure \ref{fig:graspMontage} provides a collage of static grasps generated using our contact transfer procedure, while Figure \ref{fig:fullManipulation} illustrates a manipulation sequence used to pull a box of sugar off a kitchen shelf. Note that in a number of cases the manipulator does not possess the required dexterity to successfully reach the contacts; however, the combination of our contact transfer and optimization procedure, along with ability to accommodate interactive user refinement during the process, results in kinematically-feasible ``best effort" solutions across a variety of manipulators and objects.

\subsection{Computation Speed}

To test the speed and scalability of our algorithm, we examined the time taken by each manipulator to reach a qualitatively reasonable solution across objects containing variable numbers of contact patches. Each trial of four hands was performed with respect to a single object averaged over three separate initial object placements. The same object was used for each patch set (3: scissors, 4: wineglass, 5: box, 6: lemon). All grasps had at most one contact patch per finger and one on the palm.  As a result, the three-fingered Barrett hand could not be tested for 5 or 6-contact patches, and the 4-patch Barrett hand trial was conducted using the lemon, which had a palm contact patch available. All hands were initialized to their default start positions. Parenthetical quantities indicate degrees of freedom, and no more than 6 patches were used (1 per human finger + palm). Each hand includes 6 additional degrees of freedom due to a free root joint. We model the Barrett hand's breakaway distal joints as independent degrees of freedom as a simplification due to the fact that the optimization does not model the time sequence of the grasping process. All tests were run on a single Intel Xeon W-1250 3.3 Ghz processor without the use of GPU acceleration or parallel computation.

\begin{table}[h!]
\centering
\begin{tabular}{ |>{\centering\arraybackslash}P{1.5cm}|P{1.2cm}|P{1.2cm}|P{1.2cm}|P{1.2cm}| }
    \hline
    \multicolumn{5}{|c|}{Grasp Synthesis Speed Comparison - Favorable Initialization} \\
    \hline
     & 3 patches & 4 patches & 5 patches & 6 patches\\
    \hline
    Human Hand (26) & 0.688 s / 0 edits & 0.884 s / 0.33 edits & 0.422 s / 1 edit & 1.036 s / 1 edit\\
    \hline
    Allegro Hand (22) & 0.135 s / 0 edits & 0.303 s / 0 edits & 0.278 s / 1 edit & N/A\\
    \hline
    Barrett Hand (13) & 0.081 s / 0 edits & 0.089 s / 0 edits & N/A & N/A\\
    \hline
    Prosthetic (22) & 0.288 s / 0 edits & 0.346 s / 0.33 edits & 0.284 s / 0.67 edits & 0.394 s / 1 edit\\
    \hline
\end{tabular}
\caption{Speed test comparison of 4 different manipulators of varying degrees of freedom with favorable initialization.}
\label{table:speedtestFavorable}
\end{table}

\vspace{-5mm}

\begin{table}[h!]
\centering
\begin{tabular}{ |>{\centering\arraybackslash}P{1.5cm}|P{1.2cm}|P{1.2cm}|P{1.2cm}|P{1.2cm}| }
    \hline
    \multicolumn{5}{|c|}{Grasp Synthesis Speed Comparison - Poor Initialization} \\
    \hline
     & 3 patches & 4 patches & 5 patches & 6 patches\\
    \hline
    Human Hand (26) & 1.115 s / 0 edits & 1.752 s / 0.67 edits & 1.406 s / 1.33 edits & 2.032 s / 2.33 edits\\
    \hline
    Allegro Hand (22) & 0.397 s / 0 edits & 0.721 s / 0.67 edits & 0.522 s / 1 edit & N/A\\
    \hline
    Barrett Hand (13) & 0.137 s / 0 edits & 0.183 s / 0 edits & N/A & N/A\\
    \hline
    Prosthetic (22) & 0.441 s / 0 edits & 1.443 s / 0.67 edits & 0.564 s / 1.33 edits & 1.267 s / 1.67 edits\\
    \hline
\end{tabular}
\caption{Speed test comparison of 4 different manipulators of varying degrees of freedom with poor initialization.}
\label{table:speedtestUnfavorable}
\end{table}

Tables \ref{table:speedtestFavorable} and \ref{table:speedtestUnfavorable} tabulate both the average total times taken across all optimization calls as well as the number of intermediate user edits made to $\vect{\theta}_P$ during the process. Note that these reported values only consider a small sample size and that variance between different objects and regions of contact is high; however, they do reveal some interesting trends. As expected, greater computation time and interventions are generally required for higher degree of freedom manipulators; however, it is surprising that solution search time did not necessarily increase with more contacts. A possible explanation is that additional contacts may heavily favor certain classes of solutions, and by doing so enable faster local minima convergence; however, additional experiments over larger numbers of objects and contact distributions are required to thoroughly test the conjecture.

\subsection{Robustness to Poor Initialization}

\begin{figure}
\centering
\subcaptionbox{}{\includegraphics[width=0.49\linewidth,height=0.25\linewidth]{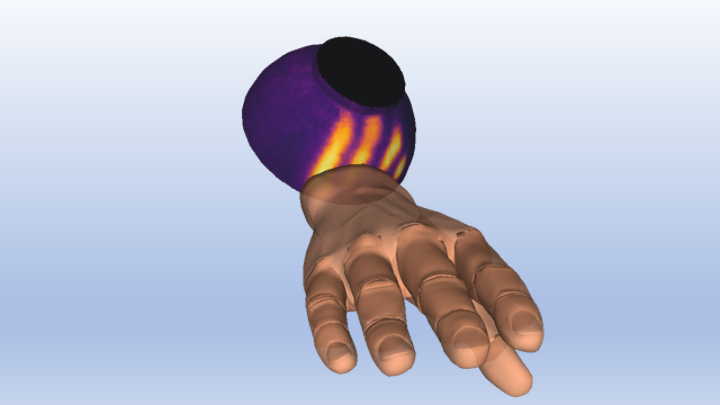}}
\subcaptionbox{}{\includegraphics[width=0.49\linewidth,height=0.25\linewidth]{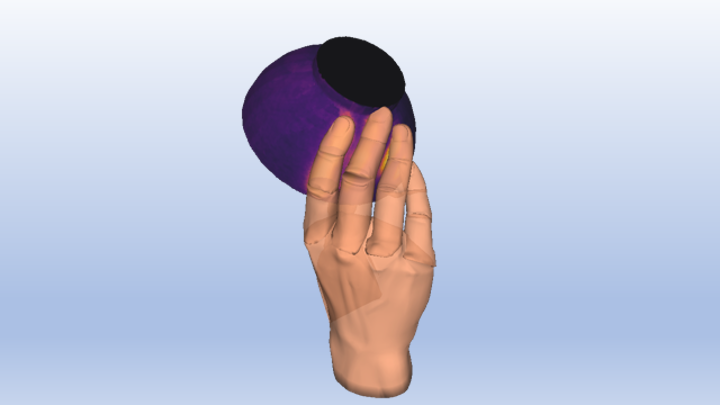}}
\caption{Results of our method in a configuration where the target object is inverted and placed behind the manipulator. Despite poor initialization (a), our method still finds an acceptable solution after only a few optimization calls (b).}
\label{fig:poorInit}
\vspace*{-0.1in}
\end{figure}

Figure \ref{fig:poorInit} shows the start state and ending grasp of the human hand after running our optimization framework for exactly two optimization calls comprised of 1,000 maximum iterations each, with each round converging and terminating in seconds with no edits. We repeated the procedure for several other poor object and hand initial states and found the behavior to be relatively consistent regardless of the manipulator and object used. As a result, our method is well suited to support dynamic object placement in a plug-and-play manner.

\subsection{Robustness to Difficult Grasps}

\begin{figure}
\centering
\subcaptionbox{}{\includegraphics[width=0.49\linewidth,height=0.25\linewidth]{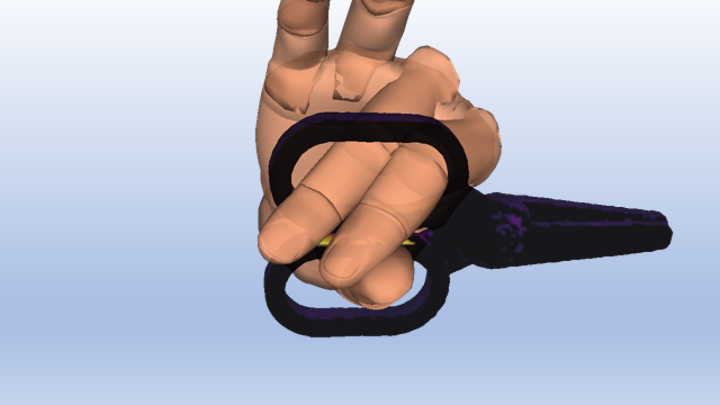}}
\subcaptionbox{}{\includegraphics[width=0.49\linewidth,height=0.25\linewidth]{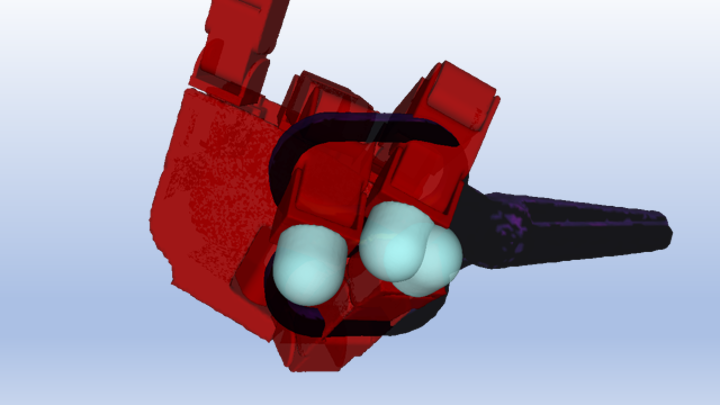}}
\subcaptionbox{}{\includegraphics[width=0.49\linewidth,height=0.25\linewidth]{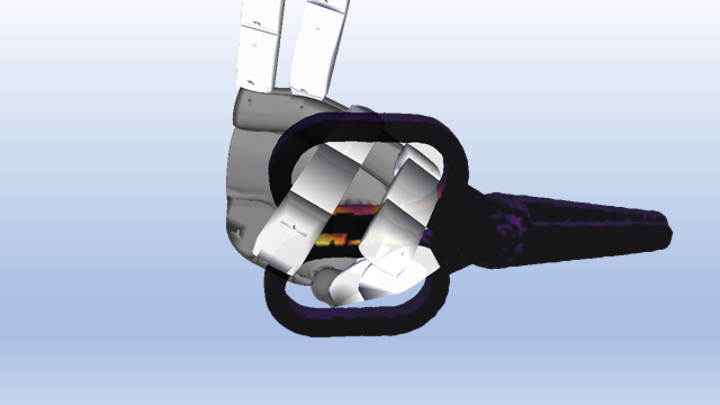}}
\subcaptionbox{}{\includegraphics[width=0.49\linewidth,height=0.25\linewidth]{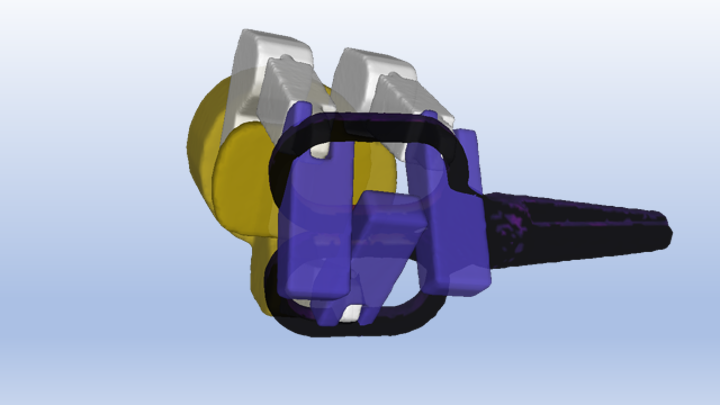}}
\caption{Results of our method in finding solutions for a difficult grasp in between the holes of a pair of scissors. The (a) human hand and (c) custom prosthetic found plausible solutions, while the (b) allegro hand and (d) barrett hand failed due to the dimensions being too large for the object.}
\label{fig:hardGrasp}
\end{figure}

Figure \ref{fig:hardGrasp} illustrates the outcome of synthesizing a kinematically-feasible grasp in between the holes of a pair of scissors. All solutions were found starting from default rest poses, and no edits were made to $\vect{\theta}_P$ between calls. While the human hand and custom prosthetic produced reasonable solutions, the Allegro and Barrett Hand noticeably appear too large to find a feasible solution.

Despite the grasp's irregularity and the fact that contacts are in hard-to-reach places, our framework still does a reasonable job in finding solutions quickly and reliably. In the case of the Barrett and Allegro Hand, the framework also provides useful information through fast failure, which can immediately prompt design changes during early stages of prototyping. Finally, due to the irregularity of the grasp and hard-to-reach contact locations, we note that grasp sampling methods would typically fail to find a solution, in contrast to our method which still manages to do so.

\subsection{Fast Reaction to Alterations}

\begin{figure}
\centering
\subcaptionbox{}{\includegraphics[width=0.49\linewidth,height=0.25\linewidth]{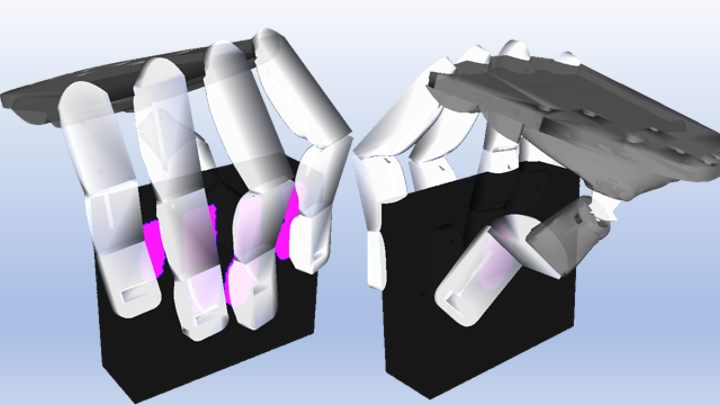}}
\subcaptionbox{}{\includegraphics[width=0.49\linewidth,height=0.25\linewidth]{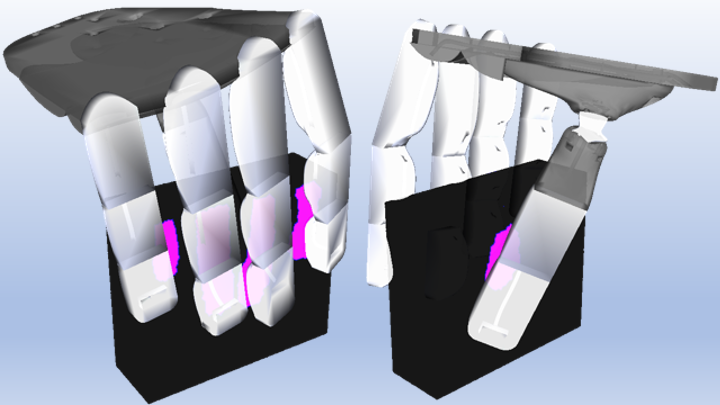}}
\caption{Custom human hand prosthesis demonstration during a box manipulation in which (a) the index finger, middle finger, and thumb joints are allowed to overextend with negative joint limits and (b) when stricter limits are imposed. Note that this alteration changes the whole hand solution rather than only the impacted joints.}
\label{fig:alterations}
\end{figure}

Figure \ref{fig:alterations} illustrates the results of running the same optimization procedure with our custom prosthesis, in which the joint limits are changed on-the-fly. In particular, we note the impact of joint limit adjustments not only on the impacted kinematic chains, but on the entire hand pose. We especially highlight the value of the latter observation within prototyping contexts since it amplifies the impact of even small design changes, which would otherwise be difficult and time-consuming to identify through existing methods.

\subsection{Robot Demonstration}

To demonstrate the utility of our approach with respect to prototyping real robot hand designs, we built a custom prosthesis using only a 3d FDM-printer, Thermoplastic Polyurethane (TPU) filament and brushless DC motors. We use a Flashforge Creator Pro 2 \cite{flashforge} retrofit with Flexion Extruders \cite{flexion} to print the hand in one single part from  soft and flexible filaments of varying shore hardness (Ninjatek Cheetah = 95A and Ninjatek Chinchilla = 75A) \cite{cheetah,chinchilla}. The fingers are actuated by tendons (monofilament nylon fishing line) that are routed through channels printed inside the hand. Joint-like kinematics are achieved either by a combination of more rigid and soft materials or by creating local geometric features such as bumps or creases on the hand. For instance, a good approximation of revolute joints is given by a crease coinciding with the joint-axis that runs across the finger or the palm. To accommodate more complex joints with more than one DoF we create rigid-soft-rigid layered features where the softer material acts as cartilage to allow for deformation along multiple axes. We place these features along the kinematic chain and investigate the potential of the hand to successfully grasp or manipulate objects. Discrepancies between simulated and real prosthesis resulting from under-actuation and soft material deformation can be reduced by selecting reasonable joint limits as shown in Figure \ref{fig:alterations}. We show that we can successfully transfer a grasp generated by our method to the real hand in Figure \ref{fig:overview} (e). Further grasps are provided in the supplemental video. 

\section{CONCLUSIONS AND FUTURE WORK}

In summary, we have presented a direct, multi-contact transfer framework that can accommodate arbitrary contact regions, objects, and manipulators. We have also presented an optimization procedure that utilizes the transferred results to produce both static grasps and manipulations quickly, reliably, and reactively to user interaction. Our method provides kinematically-feasible solutions without the use of grasp sampling or trained models, and by incorporating user feedback also enables discovery of better solutions without getting stuck in local minima. We have intentionally designed our approach to augment user capabilities, enabling full control over both the transfer and optimization process with responsive adaptation. Finally, we have shown that our method is especially useful for early stage manipulator prototyping, providing the first ground truth data-driven means of testing the impact of design parameters on the kinematics of whole hand grasps in a plug-and-play manner, while at the same time being robust to difficult to reach contacts and dynamic object placement.

However, because this paper only considers hand kinematics, it is possible that our solutions will not work in the real world. As such our next research thrust would be to extend the framework to incorporate physical  properties of the hand, object, and environment in an effort to increase the sim2real fidelity of our solutions. Additionally, enabling our algorithm to work directly on point clouds, perhaps similar that of \cite{shao2020unigrasp}, would enable greater applicability to real world environments. We are confident that the foundations laid by this paper will assist in both endeavors, and are excited to showcase the results of both thrusts in the future.



\section*{ACKNOWLEDGMENTS}

This material is based upon work supported by the AI Research Institutes program supported by NSF and USDA-NIFA under AI Institute for Resilient Agriculture, Award No. 2021-67021-35329, NSF award CMMI-1925130, and by a fellowship from CMU’s Center for Machine Learning and Health awarded to Dominik Bauer.


\newpage
\bibliographystyle{IEEEtran}
\bibliography{root}

\end{document}